\documentclass[runningheads]{llncs}

\usepackage{graphicx}
\graphicspath{{50Images/}}
\usepackage{multirow}
\usepackage{url}
\usepackage{soul}
\usepackage{amsmath}
\usepackage{color}
\usepackage[normalem]{ulem}

\begin{document}
\title{Detecting Machine-Translated~Paragraphs by Matching Similar~Words}

\authorrunning{Hoang-Quoc Nguyen-Son et al.}

\author{Hoang-Quoc Nguyen-Son \and
Tran Phuong Thao \and
\\Seira Hidano \and Shinsaku Kiyomoto}

\institute{KDDI Research Inc., Saitama, Japan\\
\email{\{ho-nguyen,th-tran,se-hidano,kiyomoto\}@kddi-research.jp}
}

\maketitle              

\begin{abstract}
Machine-translated text plays an important role in modern life by smoothing communication from various communities using different languages. 
However, unnatural translation may lead to misunderstanding, a detector is thus needed to avoid the unfortunate mistakes. 
While a previous method measured the naturalness of continuous words using a $N$-gram language model, another method matched noncontinuous words across sentences but this method ignores such words in an individual sentence. 
We have developed a method matching similar words throughout the paragraph and estimating the paragraph-level coherence, that can identify machine-translated text. 
Experiment evaluates on 2000 English human-generated and 2000 English machine-translated paragraphs from German showing that the coherence-based method achieves high performance (accuracy = 87.0\%; equal error rate = 13.0\%). 
It is efficiently better than previous methods (best accuracy = 72.4\%; equal error rate = 29.7\%). 
Similar experiments on Dutch and Japanese obtain 89.2\% and 97.9\% accuracy, respectively.
The results demonstrate the persistence of the proposed method in various languages with different resource levels.

\end{abstract}

\section{Introduction}

Machine translation is the most vital assistance in communication between two persons comprehending different languages, so renowned international companies such as Facebook and Google integrate translators into text content including blogs, web-pages, and comments.
While translation is increasingly developed for rich resource languages, especially in European community having a strong connection in economy and culture; the low resource languages such as Arabic, Pashto, and Dari are also initially investigated to prevent potential risks from criminal actions\footnote{\url{https://www.pri.org/stories/2011-04-26/machine-translation-military}}, for example, terrorism and kidnapping.

\begin{figure}[t]
\centering
\includegraphics[]{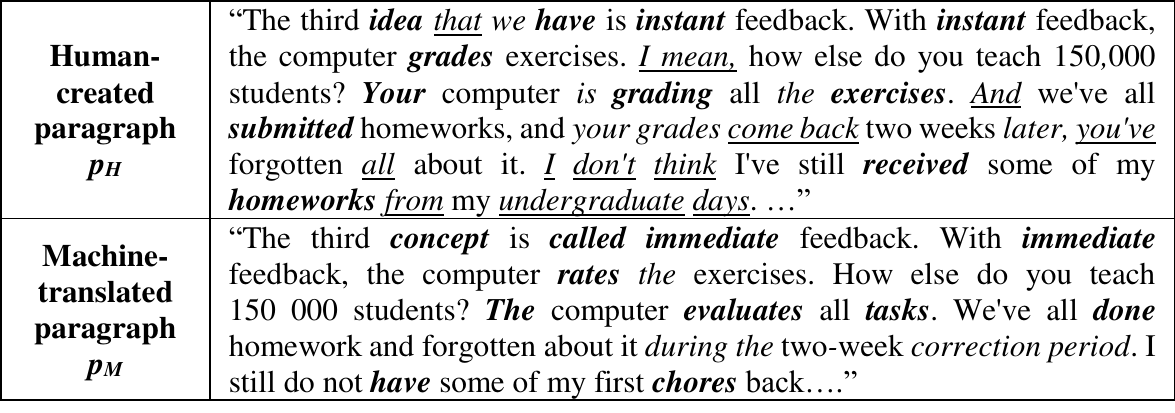}
\caption{Coherence of human-created vs machine-generated paragraph.}
\label{fig:01_CoherenceExamples}
\end{figure}

Although a machine preserves meaning in a translated text, the use of `strange' words reduces readability.
Figure~\ref{fig:01_CoherenceExamples} illustrates different quality of original and translation\footnote{\url{https://www.ted.com/talks/anant_agarwal_why_massively_open_online_courses_still_matter/transcript}} despite same content in each.
A machine can correctly generate grammatical text, but the selection of vague words may result in misunderstanding, especially in the last sentence of the figure.
The confusing can be reduced by recognizing and notifying translated text to readers.

Many methods have been published different approaches to detecting translation text.
These approaches can be categorized by core techniques: parsing tree, $N$-gram model, word distribution, and word similarity.
The first approach extracted distinguishable features from \textit{parsing trees}~\cite{chae2009predicting,li2015machine}, but such trees are only parsed from an individual sentence. 
To overcome this problem, other methods~\cite{aharoni2014automatic,arase2013machine,quoc2017detecting} based on \textit{$N$-gram language model} extract such features from nearby words in both inside and outside a sentence. 
The limitation of this model is that meaningful features are only given from few nearby words, common in three.
Other work~\cite{labbe2013duplicate,nguyen2017identifying} analyzes the histogram of \textit{word distribution} from a massive amount of words, particularly suitable for document level.
A recent method~\cite{nguyen2018identifying}, the closest one to this paper, estimates text coherence by mutually matching words across pairwise sentences using \textit{word similarity}.
This method, however, ignores the connection in such words within a sentence.

The coherence of a human-generated paragraph is often higher than that of a machine-translated one. 
In Figure~\ref{fig:01_CoherenceExamples}, for instance, the high coherence makes the human-generated text more comprehensible.
We analyze coherence by highlighting the difference of word usage in italic.
A machine commonly uses quite different words, emphasized in bold, that affect the preservation of intrinsic meaning.
For example, in the third sentence, ``\textit{tasks}'' replaces ``\textit{exercises}'' easily leading to misunderstanding.
Moreover, the translation misses some subordinate words that are marked in underline in the man-made paragraph.
According to Volanskey et al.~\cite{volansky2013features}, such words significantly improve text comprehension.
Thus, the missing of the certain words, especially in the last sentence, makes the translated text more confusable.

In this paper, we have proposed a method for matching similar words in a paragraph with maximum similarity.
The similarity is then used to estimate coherence that can determine whether a paragraph is translated by a machine or created by a human.

We collected TED talk\footnote{\url{https://www.ted.com/}} transcripts written by native speakers and chose only transcripts aligned with both German and English.
While English represents for human-generated text, the German is translated into English by Google to produce machine-generated one.
The best translator, Google, can create not only the highest quality translation but also the most difficult to distinguish, as demonstrated in Aharoni et at.'s work~\cite{aharoni2014automatic}.
We then randomly select 2000 paragraph pairs for conducting experiments.
The results show that the coherence-based method accomplishes superior accuracy 87.0\% and low equal error rate 13.0\%.
It surpasses previous methods with the best accuracy 72.4\% and equal error rate 29.7\%.

The coherence-based method also reaches the highest performances on Dutch and Japanese with similar experiments.
It demonstrates the persistence of the proposed method in various languages.
While Dutch has competitive results with German, Japanese even obtains impressive accuracy 97.9\% and mere equal error rate 1.9\%.
It indicates lack of coherence on the low resource language.
Based on this finding, translators can enhance text coherence by enriching linguistic resources.

In the rest of this paper, Section~\ref{sec02:related-work} outlines the main previous methods of machine-translated text detection. 
Section~3 describes a step-by-step guide to extract coherence features.
Experiments on these features and comparison of the coherence-based with previous methods on various languages are shown and analyzed in Section~4. 
Finally, Section~5 summaries some main key points and mentions future work.

\section{Related Work}\label{sec02:related-work}

Since machine-translated text detection is an important task of natural language processing, many researchers have involved suggesting useful solutions.
The previous solutions can be grouped by the core usage including parsing tree, $N$-gram model, word distribution, and word similarity. 
Some main methods of each group are summarized in below.

\subsection{Parsing Tree}

In this approach, researchers aimed to extract detectable features from a parsing tree for use in machine-translated text identification.
For example, Chae and Nenova~\cite{chae2009predicting} claimed that parsing of machine text is commonly simpler than that of human one. 
The authors indicated that a simple parsing often contains short main constituents, that is noun, verb, and adjective phrases.
Following the intuition, the authors extracted meaningful features, such as parsing tree depth, phrase type proportion, average phrase length, phrase type rate, and phrase length, before using them to distinguish computer- with human-generated text.

Li et al.~\cite{li2015machine} inherited several above features including parsing tree depth and phrase type proportion.
In addition, they investigated that the structure of human parsing is more balancing than that of machine one. 
They thus suggested some useful features: the ratio of right- compared to left-branching nodes, the number of left-branching nodes for noun phrases. 
The main limitation of parsing-based methods is that they just generate a parsing tree for an individual sentence. 
An integrated tree cannot be built for larger scope of multiple sentences such as paragraph or document.

\subsection{$N$-gram Model}

To overcome the limitation of parsing-based approach, Arase and Zhou recommended another method~\cite{arase2013machine} based on fluency estimation. 
They mainly used $N$-gram language model to estimate the fluency of continuous words.
The restriction of this model is that it efficiently examines only on few continuous words, common in three.
The authors reduced the deficiency by using sequential pattern mining to measure the fluency of non-continuous words.
In-fluent patterns in human text are mined, such as ``\textit{not only * but also},''`` \textit{more * than},'' that contrast with that in machine-generated text, for example, ``\textit{after * after the},'' ``\textit{and also * and}.''
There are two other reasonable combinations also aim to diminish the restriction of $N$-gram model.
The first combination~\cite{quoc2017detecting} extracted the specific noise words often used by a human, that is misspelled and reduction words, or by a machine, namely untranslated words.
This combination, however, is only efficient in online social network in which contains a substantial number of such noises.
The second combination~\cite{aharoni2014automatic} focused on functional words abundantly occurring in machine-translated text. 
Additional features in the three combinations achieve non-high performances but these features effectively improve the overall performances when they are integrated with the original $N$-gram model.
 
\subsection{Word Distribution}

Another approach recognizes machine-generated text by analyzing a histogram of word distribution.
For example, Labb{\'e} and Labb{\'e} suggested an inter-textual metric for estimating the similarity of word distributions~\cite{labbe2013duplicate}.
This metric is perfectly used for classifying artificial and real papers with accuracy up to 100\%, but Nguyen-Son et at.~\cite{nguyen2017identifying} indicated that the inter-textual metric is just suitable for paper detection and they developed another method for translation detection also based on word distribution.
This method pointed out that a word distribution of human text is closer with a Zipfian distribution than that of machine one. 
They also offered some valuable features to support the word distribution, that is specific phrases (e.g., idiom, clich{\'e}, ancient, dialect, phrasal verb) and co-reference resolution.
The restriction of distribution-based methods is that they are only stable with a large number of words.
However, the deficiency is revealed on homologous texts that refers to same sources such as paraphrasing and translation because such text mostly contains a same set of words.

\subsection{Word Similarity}

The closest method with our work was suggested by Nguyen-Son et al.~\cite{nguyen2018identifying}.
They matched similar words in pairwise sentences of a paragraph.
In two sentences, each word is only matched with another word at most so that total similarity of matching is maximum.
We extend this idea by matching similar words in both internal and external sentences throughout a paragraph, so a word can be used as a bridge of other words in the text. 

\section{Proposed Method}

\subsection{Overview}

\begin{figure}[t]
\centering
\includegraphics[]{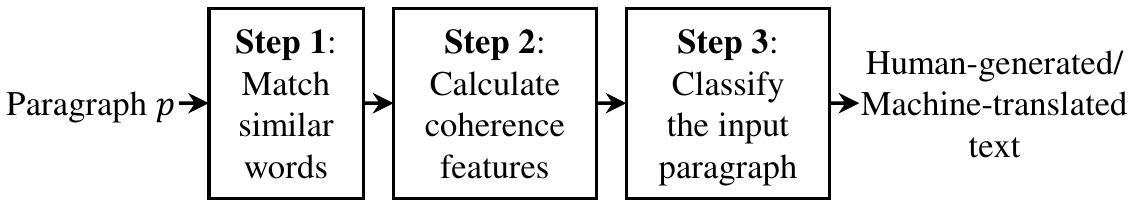}
\caption{Proposed schema for computer-generated paragraph detection.}
\label{fig:02_proposed_schema}
\end{figure}

The proposed schema distinguishes between machine-translated and human-generated paragraphs in three steps shown in Figure~\ref{fig:02_proposed_schema}.

\begin{itemize}
\item
\textbf{Step 1 (\textit{Match similar words})}: 
Each word is matched with other words in the input paragraph $p$. 
The similarity of matched pairs is measured by Euclidean distance.
The maximum similarity is distributed into disparity groups based on part of speech of the matched words.

\item
\textbf{Step 2 (\textit{Calculate coherence features})}: 
Mean and variance metrics are calculated for all similarity in each group.
These metrics are used as features to estimate the coherence of $p$.

\item
\textbf{Step 3 (\textit{Classify the input paragraph})}: 
The coherence features are used to determine whether the input $p$ is created by a human or is translated by a machine.
\end{itemize}

The three-step is presented in detail and demonstrated by the human-created paragraph $p_H$ and the machine-translated one $p_M$ in Figure~\ref{fig:01_CoherenceExamples}.

\subsection{Detail}

\subsubsection{Matching similar words (Step 1)}

\begin{figure}[t]
\centering
\includegraphics[]{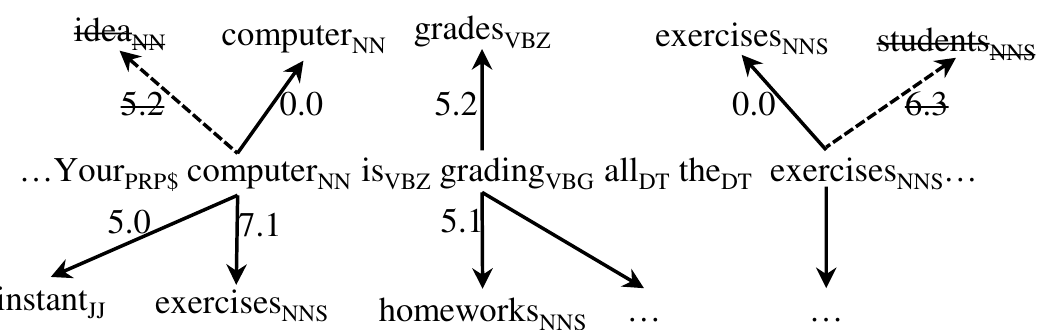}
\caption{Matching main words in human-generated paragraph $p_H$.}
\label{fig:03_word_matching_human}
\end{figure}

\begin{figure}[t]
\centering
\includegraphics[]{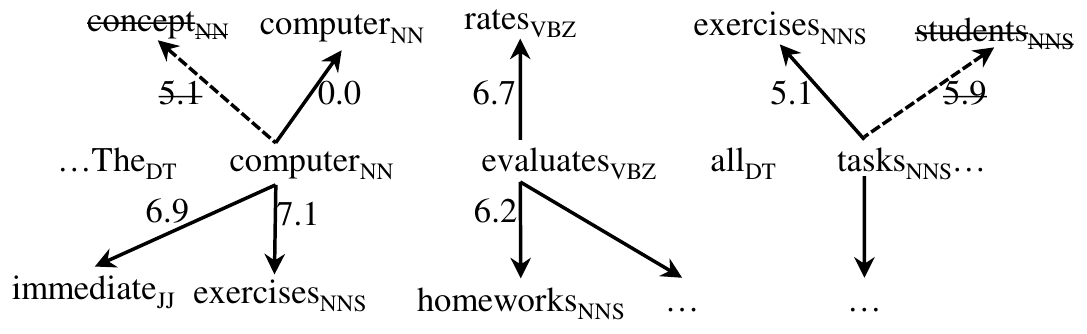}
\caption{Matching main words in machine-translated paragraph $p_M$.}
\label{fig:04_word_matching_machine}
\end{figure}

Words in the input paragraph $p$ are separated and labeled with parts of speech (POS) using Stanford tagger~\cite{manning2014coreNLP}.
A word is then matched to other words, and each similarity of two matched words is measured.
The similarity is estimated by the distance of two vectors on a word embedding. 
We use a common word embedding, GloVe~\cite{pennington2014glove}, that is trained from Wikipedia 2014 merging with Gigaword 5 and produces 400K vectors with 300 dimensions in each. 
The Euclidean is chosen here due to much wider distance comparing with Cosine.
In Euclidean space, the distance of vectors is larger, so the difference of words is clearer.
In there, the higher similarity of two words indicates the lower value of the distance.
Some of matched pairs are plotted in Figure~\ref{fig:03_word_matching_human} and Figure~\ref{fig:04_word_matching_machine} for the human paragraph $p_H$ and the machine one $p_M$, respectively.

If a word is matched with other words having the same POSs, then the minimum distance is preserved.  
In Figure~\ref{fig:03_word_matching_human}, for example, a singular noun ``\textit{computer}''(NN) can be matched with two singular nouns, namely ``\textit{idea}'' and ``\textit{computer},'' having the similarities 5.2 and 0.0, respectively. 
The lower distance ``\textit{computer-computer}''(0.0) is chosen while the other matching is eliminated that is marked in strike-through.

As shown in the two figures, the similarity of a human-generated text tends lower than that of a machine-translated one. 
It demonstrates the high coherence of human-created passages, on the other hand, the use of low-coherent words reduces the overall coherence of the machine-translated text. 
For example, a pair ``\textit{evaluates-rates}'' (6.7) in $p_M$ causes to slightly drop the coherence when it is compared to a higher coherence pair ``\textit{grading-grades}'' (5.2) in $p_H$.
The difference also affects to other matching such as ``\textit{grading-homeworks}'' versus ``\textit{evaluates-homeworks}.''
Similar cases occur in other pairs, for instance, \{``\textit{exercises-exercises},'' ``\textit{computer-instance}''\} versus \{``\textit{exercises-tasks},''``\textit{computer-immediate}''\} in human versus machine text, correspondingly. 
It is easy to confuse readers who possibly understand the meaning in various ways.

\subsubsection{Calculating coherence features (Step 2)}

\begin{figure}[t]
\centering
\includegraphics[]{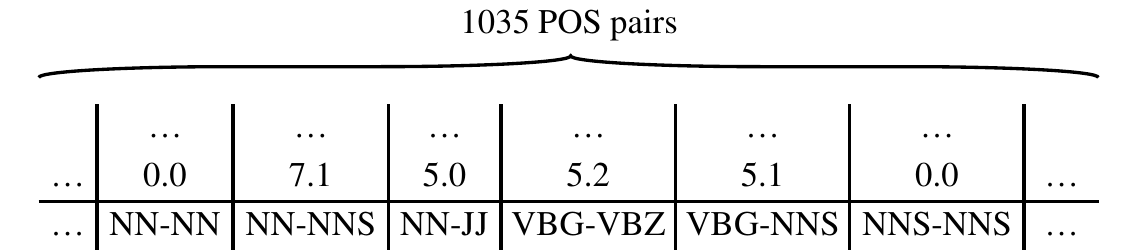}
\caption{Distributing similarities to part of speech (POS) groups.}
\label{fig:05_similarity}
\end{figure}

The similarity of the remaining pairs is distributed to certain groups based on their POSs.
For example, while ``\textit{computer-computer}'' (0.0) in $p_H$ is allocated to NN-NN group, ``\textit{computer-exercises}'' (7.1) is delivered to NN-NNS as shown in Figure~\ref{fig:05_similarity}.
The number of groups equals 1035 created from a list of 45 separate POSs.

Means and variances are calculated in each group for estimating the text coherence.
2070 values including 1035 means and 1035 variances are used to detect machine-translated paragraphs in the next step.

\subsubsection{Classifying the input paragraphs (Step 3)}

The statistical values represent as coherence features to determine whether the input $p$ is a computer- or a human-generated paragraph. 
These features are examined on three common machine learning classifiers that were chosen in state-of-the-art methods including linear classification~\cite{fan2008liblinear}, support vector machine (SVM) optimized by stochastic gradient descent (SGD), SVM optimized by sequential minimal optimization (SMO). 
The SVM(SMO) reaches the best performance, so it is selected as the final classifier.

\section{Evaluation}
\subsection{Dataset}

We collected 3088 English and 2253 German transcripts, that are composed by native speakers in TED talks and posted from June 2009 to November 2018.
We then choose the transcripts existing in both English and German and aligned in paragraph-by-paragraph.
While the rest English text is considered as a human creation, the German is translated into English by Google for generating machine-generated paragraphs.
Finally, we randomly selected 2000 aligned pairs to conduct experiments. 
Each paragraph contains 14.4 sentences in average.

\subsection{Comparison}

The dataset is evaluated by 10-fold cross-validation on three common machine learning algorithms including linear classification (LINEAR), support vector machine (SVM) optimized by stochastic gradient descent SVM(SGD), or by sequential optimization SVM(SMO). 
These classifiers reached best performances on previous machine translation detection methods. 
Since $F$-measure and accuracy (ACC) are analogous to results, the only ACC is shown in this experiment.
We also calculate equal error rate (EER) to test the persistence of each classifier.
The coherence-based method using mean, variance, and their combination in both Cosine and Euclidean distance is compared with four previous methods on the same task.
While three methods based on word distribution with coreference resolution (coreref)~\cite{nguyen2017identifying}, $N$-gram model~\cite{aharoni2014automatic}, and word similarity~\cite{nguyen2018identifying} can directly extract features from a paragraph, the  other~\cite{li2015machine} based on parsing tree only obtains such features from an individual sentence.
Thus, we adopt this method for a paragraph by calculating average on the features. 
The results of the comparison are shown in Table~\ref{tab:01Comparison}.

\begin{table}[t]
\begin{center}
\caption{Comparison with previous methods on accuracy (ACC) and equal error rate (ERR) metrics. The underline describes for the best classifiers, which are selected in the previous methods. The best values of each work are emphasized in bold, and the highest performance among them is highlighted by red.}
\label{tab:01Comparison}
\begin{tabular}{l l c c c c c c}

\hline 
\multicolumn{2}{c }{\multirow{2}{*}{\textbf{Method}}} &\multicolumn{2}{c}{\textbf{LINEAR}} & \multicolumn{2}{ c }{\textbf{SGD(SVM)}} & \multicolumn{2}{c}{\textbf{SMO(SVM)}}  \\ 
\multicolumn{2}{c }{} &\textbf{ACC} &\textbf{EER} &\textbf{ACC}&\textbf{EER} &\textbf{ACC}&\textbf{EER}\\ 
\hline
\multicolumn{2}{l}{Word distribution and coreref~\cite{nguyen2017identifying}}	&	66.5\%    &    33.4\%  &    \underline{66.6\%}    &    \underline{\textbf{33.3\%}}    &    \textbf{66.9\%}    &    33.3\%    \\
\multicolumn{2}{l}{Parsing tree~\cite{li2015machine}} &\underline{\textbf{67.9}\%}    &     \underline{33.4\%} &   67.0\%    & 34.4\%    &    67.6\%    &    \textbf{32.8\%}      \\
\multicolumn{2}{l}{$N$-gram and functional words~\cite{aharoni2014automatic}} &	\textbf{69.5\%}    &    \textbf{30.5\%}  &    67.0\%    &    32.9\%    &    \underline{69.3\%}    &    \underline{30.8\%}    \\
\multicolumn{2}{l}{Word similarity~\cite{nguyen2018identifying}}&	  \underline{\textbf{72.4\%}}    &    \underline{\textbf{29.7\%}}&    69.6\%    &    31.1\%    &    70.9\%    &    30.8\%  \\
\hline 
{\multirow{3}{*}{Cosine}}&Mean &	83.8\%    &    15.5\%    &    80.0\%    &    23.5\%    &    \textbf{84.6\%}    &    \textbf{15.6\%}  \\
& Variance &	67.8\%    &    32.5\%    &    70.3\%    &    31.1\%    &    \textbf{72.8\%}    &    \textbf{27.3\%}  \\
& Combination &	83.9\%    &    17.0\%    &    81.1\%    &    21.3\%    &    \textbf{85.4\%}    &    \textbf{14.6\%}  \\
\hline 
{\multirow{3}{*}{Euclidean}}& Mean &	83.2\%    &    19.0\% &    83.3\%    &    18.5\%    &    \textbf{85.6\%}    &    \textbf{14.0\%}  \\
& Variance&	75.7\%    &    21.4\%  &    76.3\%    &    25.3\%    &    \textbf{79.5\%}    &    \textbf{20.7\%}    \\
& Combination&	84.1\%    &    16.8\% &    84.8\%    &    15.4\%    &    {\color{red}\textbf{87.0\%}}    &    {\color{red}\textbf{13.0\%}}      \\

\hline 

\end{tabular}
\end{center}
\end{table}

The accuracy in Table~\ref{tab:01Comparison} is in harmony with EER that demonstrates the high persistence through classifiers.
Through all previous methods, the highest performances (in bold) are identical or competitive to best classifiers indicated in italic with the maximum deviation only 0.6\%.
In these classifiers, the method based on word distribution and coreference resolution (coreref)~\cite{nguyen2017identifying} attained the lowest performance.
The main reason is that the distribution is affected by a limited number of words within a paragraph.
The parsing-based method~\cite{li2015machine} slightly improves the performance.
However, the parsing can only be built from an individual sentence, so the relationship of words in cross-sentence is ignored.
Another method~\cite{aharoni2014automatic} based on $N$-gram model and functional words are more suitable for paragraphs but this model is just efficient on few consecutive words. 
On the other hand, a similarity-based method~\cite{nguyen2018identifying} exploits the extra connections among nonconsecutive words and accomplishes the current state-of-the-art performance (accuracy = 72.4\%; EER = 29.7\%).

The coherence-based method achieves the superior performances comparing with previous work through all classifiers.
The Euclidean distance brings higher results due to the large diversity in measuring word similarity. 
Although mean is more appropriate to evaluate the text coherence than variance, the later one significantly supports to enhance the overall outcomes.
Therefore, the combination archives topmost performances.
The best performance (accuracy = 87.0\%; EER = 13.0\%) is obtained when using SMO(SVM) classifier. 
This classifier is thus chosen for further experiments on the coherence-based method while the other methods are evaluated on the best classifiers chosen in corresponding papers (underline in Table~\ref{tab:01Comparison}).

\subsection{Individual Features}

We examine the coherence-based method on each individual feature with the experimental results shown in Table~\ref{tab:02BestPOSPairs}.
The performances are sorted by combination accuracy for finding most important features contributing to estimate the coherence.
The outcomes demonstrate the mutual support between mean and variance in the combinations.
In top pairs, the colon indicates the pivot role when it is combined with other POSs.
Since the `:' rarely occurs in a machine-translated text, this mark significantly provides for recognizing the artificial translation.
Because the mark is often used to explanation, the missing of colon causes reducing the clarity of the translated text comparing with the original version.
The translators can upgrade the text coherence by integrating this mark into the translation.

\begin{table}
\caption{Performances of top five POS pairs.}
\label{tab:02BestPOSPairs}
\begin{center}
\begin{tabular}{l l l c c c }

\hline 
\textbf{Rank}& \textbf{POS pair  } & \textbf{Mean} & \textbf{Variance}& \textbf{Combination}\\ 
\hline
1 & TO-: & 48.3\% & 60.5\%  & 72.0\%\\
2 & VBP-RB & 61.2\% & 55.7\% & 68.1\% \\
3 & :-WP & 60.9\% & 56.3\% & 66.6\% \\
4 & WRB-: & 55.8\% & 59.1\% & 66.5\% \\
5 & PRP-RB & 63.2\% & 49.8\% & 66.2\% \\
\hline
\end{tabular}
\end{center}
\end{table}

\subsection{Other Languages}

\begin{figure}[t]
\centering
\includegraphics[width=\textwidth]{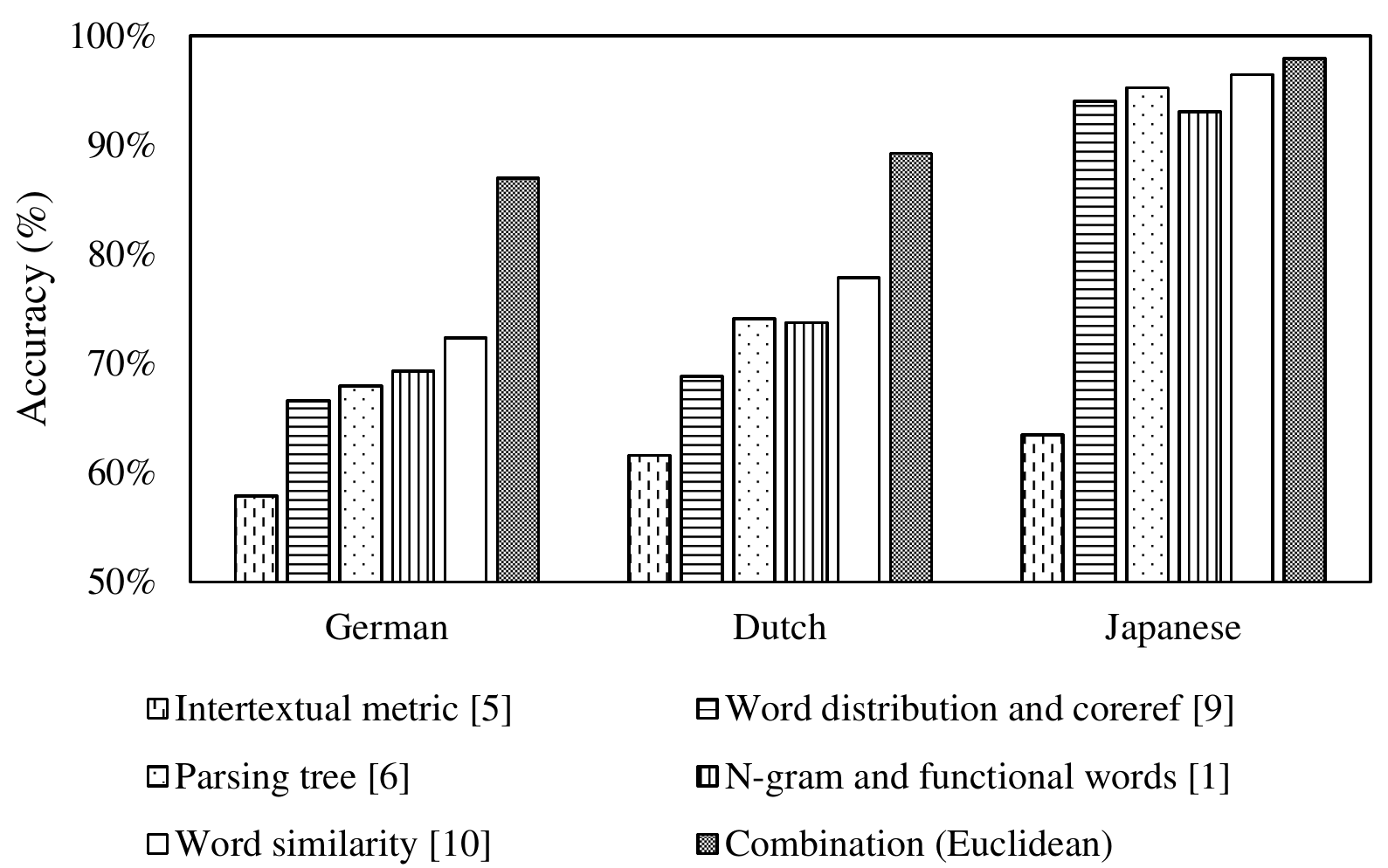}
\caption{Evaluation on various languages.}
\label{fig:06_OtherLanguages}
\end{figure}

We conduct similar experiments with two other languages including Dutch and Japanese.
While Dutch is another rich resource like German, the Japanese is a low resource language.
2000 English paragraphs are chosen in each language for human-generated text.
The aligned 2000 Dutch and 2000 Japanese paragraphs are translated into English by Google for producing machine text.
The results of the experiments are plotted in Figure~\ref{fig:06_OtherLanguages}.
Because equal error rate associates with accuracy, the only accuracy is shown in the chart.
We also evaluate on another method~\cite{labbe2013duplicate} based on word distribution that is recommended by Labb{\'e} and Labb{\'e}.
While other methods extract features to run on machine learning classifiers, the distribution-based method suggested an inter-textual distance to measure the distance between two distributions.

In Figure~\ref{fig:06_OtherLanguages}, although the inter-textual metric impressively recognizes an artificial document, it is insufficient to apply for lower granularity such as a paragraph.
In other words, the method is degraded by a limited amount of words in a text.
Therefore, this work archives almost same performances due to similar word distribution on these languages.
On the other hand, other methods measure the text fluency, so the performances are obviously changed between low and rich resource languages.
In remaining methods, while the performances on Dutch are similar to German, evaluation on Japanese archives notable improvement.
It demonstrates the significant impact of resource level on the fluency measurement.
Moreover, the grammar structure is also another important aspect.
English uses a common structure SVO, i.e. a subject follows by a verb and an objective, but most of Japanese sentences are SOV. 
In previous methods, the approach~\cite{nguyen2018identifying} based on word similarity is the most stable and reaches higher performances.
 
\begin{figure}[t]
\centering
\includegraphics[]{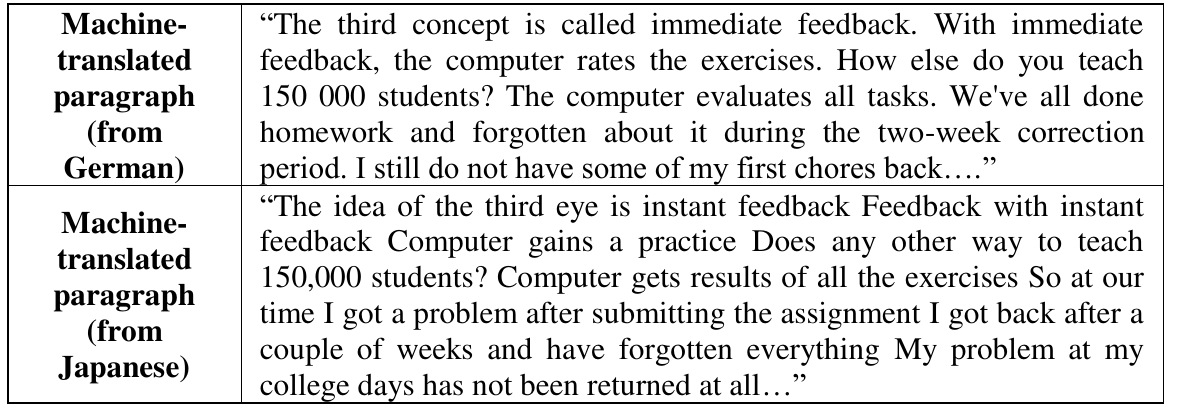}
\caption{The machine-translated text from German and Japanese.}
\label{fig:07_Japanese_German}
\end{figure}

The coherence-based method outperforms other methods in all three languages.
While Dutch is similar to German, Japanese clearly improves the accuracy up to 97.9\%.
It indicates the poor coherence of machine-translated text in the low-resource language. 
Another translation of the text in Figure~\ref{fig:01_CoherenceExamples} is translated from Japanese as shown in Figure~\ref{fig:07_Japanese_German}.
Comparing to German, the Japanese version is obviously lower quality, so it leads to hard-understand the intrinsic meaning.
The finding can be used to justify the quality of a machine translator.
Based on that, the translator can improve the text coherence by enriching resources in such languages.

\section{Conclusion}

We propose a method for identifying machine-translated paragraph using coherence features.
Each word is matched to other words through a paragraph with the maximum similarity.
The similarity represents for text coherence and is used to distinguish human-created with machine-translated paragraphs.
Experiments on German show that the coherence-based method archives superior performance (accuracy = 87.0\%; equal error rate = 13.0\%) when it is compared with other state-of-the-art methods.
Similar experiments on Dutch obtain equivalent results while evaluation on Japanese reaches superb accuracy 97.9\% and mere equal error rate 1.9\%. 
The results indicate that text coherence is affected by a resource level.
The coherence can also be used to measure the quality of a machine translator.

The current work focuses on classifying human-written and machine-translated text. 
In future work, we aim to produce and evaluate man-made and artificial translation.
We also target on estimating the coherence on a website for referring readability.
Moreover, a deep learning network can be used to enhance the coherence measurement.
 
\bibliographystyle{splncs04}      
\bibliography{CICLING_2019}
\end{document}